# Real-Time Atomic-Resolution Electron Phase Imaging without Probe Calibration via Ptychography-Supervised Learning

*H. Yue[1], C.-C. Chen[2, *], C.-N. Hsiao[3], J. Cheng[4], Y. Liu[5], X.Z. Liao[5] and Steve F. Shu[6]*
[1]*College of Intelligent Robotics and Advanced Manufacturing, Fudan University, Shanghai 200433, China.*
[2]*Department of Engineering and System Science, National Tsing Hua University, Hsinchu 300044, Taiwan*
[3]*National Center for Instrumentation Research, National Institutes of Applied Research, Hsinchu 300092, Taiwan*
[4]*Institute for Infocomm Research, A*STAR, Singapore, 138632, Singapore*
[5]*School of Aerospace, Mechanical and Mechatronic Engineering, and Australian Centre for Microscopy and Microanalysis, The University of Sydney, Sydney, NSW 2006, Australia.*
[6]*School of Electrical and Computer Engineering, The University of Sydney, Sydney, NSW 2006, Australia*
[*]Corresponding Author: cheinchunchen@mx.nthu.edu.tw

KEYWORDS: Electron ptychography; 4D-STEM; phase imaging; real-time microscopy; ptychography-supervised learning

ABSTRACT**:** Atomic-scale phase imaging is central to resolving defects, interfaces, and weakly scattering atoms that govern the behavior of nanoscale materials. Electron ptychography delivers sub-ångström phase sensitivity but remains an offline technique, because its iterative reconstruction is computationally expensive and sensitive to experimental calibration, preventing live use during data acquisition. Here, a ptychography-supervised local inference framework is presented that converts four-dimensional scanning transmission electron microscopy (4D-STEM) into an acquisition-compatible phase-imaging workflow. Physics-constrained reference phase maps reconstructed from a single experimental AuPd dataset serve as teacher labels for a compact model that predicts local phase patches directly from diffraction measurements, without explicit probe input or online iterative optimization. Full-field images are assembled by deterministic overlap stitching. The workflow reaches an online latency of about 0.27 ms per probe position and a throughput of about 20,000 positions per second, an approximately 1,000-fold speed-up over GPU-accelerated ePIE, while preserving atomic-scale lattice contrast and reciprocal-space fidelity. Without fine-tuning, the same model transfers across materials ($WS_2$), defocus conditions (high-entropy alloy nanoparticles), and instruments (hBN at 300 kV). The approach amortizes ptychographic redundancy into a fast, generalizable workflow that enables real-time atomic-scale phase imaging for materials microscopy.

## Introduction

Atomic-scale imaging is central to understanding how local structure, defects, interfaces, and weakly scattering atoms control the behavior of nanoscale materials. In scanning transmission electron microscopy (STEM), annular dark-field imaging provides direct structural contrast for many crystalline materials, but it is less sensitive to weak phase variations and light-element contrast. Four-dimensional STEM (4D-STEM), in which a convergent-beam electron diffraction pattern is recorded at each probe position by a pixelated detector, provides a richer measurement of the specimen-probe interaction.[1,2] Electron ptychography uses this four-dimensional measurement to recover quantitative phase information, enabling sub-angstrom phase contrast and structural analysis beyond conventional image-formation limits.[3-6]

Despite this promise, electron ptychography remains difficult to use as a live imaging method during experiments. Conventional reconstruction uses highly overlapping diffraction measurements to constrain the phase-retrieval problem and enforce consistency between the object and probe.[7-9] Algorithms such as the extended ptychographic iterative engine (ePIE) therefore solve a globally coupled inverse problem over many probe positions.[9] This process is physically grounded, but it is computationally expensive and can require careful treatment of reconstruction parameters, probe behavior, scan-coordinate consistency, defocus, and other microscope-dependent factors.[11,26-28] As a result, ptychographic phase maps are usually reconstructed after data acquisition rather than during scanning. This delay limits the use of phase contrast for real-time decisions, including field-of-view selection, drift assessment, dose control, and rapid screening of heterogeneous materials.

Deep learning offers a possible route to faster ptychographic imaging and phase recovery.[12-17] However, direct supervised reconstruction also introduces important risks. A network trained to map diffraction measurements to phase images may learn correlations that are specific to one probe, defocus condition, material system, or microscope. This is particularly serious in electron ptychography because the measured diffraction signal contains both intrinsic specimen information and extrinsic illumination effects. A model that performs well under matched conditions can therefore degrade when the probe shape, defocus, detector response, or sample class changes, consistent with known instability and generalization concerns in deep-learning image reconstruction.[18,19] For real experimental use, acceleration alone is not sufficient; the reconstruction workflow must retain the physical constraints that make ptychography reliable while reducing the need for online iterative optimization.

Here, a ptychography-supervised local inference framework is introduced to convert high-overlap electron ptychography from an offline reconstruction method into an acquisition-compatible phase-imaging workflow. The central idea is not to replace ptychography with an unconstrained single-frame neural network. Instead, the physical redundancy used by ePIE is first used offline to generate coordinate-consistent teacher phase maps from an experimental 4D-STEM dataset. A compact neural model is then trained to infer local ePIE-consistent phase patches directly from convergent-beam electron diffraction measurements within the probe coherent core. During inference, the model predicts local phase patches without explicit probe input or online global optimization. Full-field phase images are assembled by deterministic overlap stitching using the relative scan coordinates.

This design shifts the role of ptychographic redundancy. In iterative reconstruction, overlap, probe relaxation, and object consistency are enforced online for every new dataset through a coupled inverse problem. In the present workflow, these constraints are first embedded into teacher reconstructions and then amortized into a local phase estimator. The remaining role of overlap during deployment is to support geometric assembly and boundary consistency, rather than to drive iterative phase retrieval. The output should therefore be understood as a fast, experimentally grounded, ePIE-consistent phase estimate, with validity set by the teacher reconstruction, sampling density, probe shift, dose, and scattering regime.

The method is trained using ePIE-derived phase patches from a single experimental AuPd 4D-STEM dataset. AuPd is used as the training source because its polycrystalline nanoparticle structure provides diverse local orientations, grain boundaries, and short-range phase variations. The same trained model is then applied without fine-tuning to several out-of-distribution conditions, including spatially held-out AuPd, WS2, a defocus series of high-entropy alloy nanoparticles, and cross-instrument hBN data. These tests examine whether a local phase prior learned from one structurally rich experimental source can transfer across material, defocus, probe, and instrument changes.

The resulting workflow reaches an online latency of approximately 0.27 ms per probe position, including preprocessing, inference, and stitching, and supports batched processing at about 20,000 probe positions per second. Across the tested datasets, the model preserves lattice contrast and reciprocal-space peak structure relative to ePIE reference reconstructions, while also revealing the limits imposed by sparse sampling, large probe-domain shifts, and low-dose noise. These results show that ptychographic redundancy can be amortized into a fast local inference model, enabling real-time electron phase imaging for experimental materials microscopy while retaining a clear connection to physics-constrained ptychographic reconstruction.

## Results

### Ptychography-supervised local inference enables real-time streaming phase imaging

We first evaluated whether the proposed ptychography-supervised local inference framework can match the acquisition rate of experimental 4D-STEM (Fig. 1). In the conventional workflow, high-overlap 4D-STEM data are acquired as a full datacube and reconstructed by iterative phase retrieval, such as ePIE, to obtain coordinate-consistent ptychographic reference phase maps (Fig. 1a,b). In our workflow, this iterative reconstruction is used offline for teacher generation and local training, whereas deployment is performed by one-pass local phase inference followed by deterministic stitching using the relative scan coordinates (Fig. 1c,d). This changes where the main computation occurs: the physically constrained ptychographic reconstruction is used during training, while online imaging requires only local prediction and field assembly.

Under a standard 220 μs detector dwell time, corresponding to an effective sampling interval of approximately 0.50 ms per frame after including I/O and binning, the total online processing latency is approximately 0.27 ms per frame (Fig. 1d). This includes preprocessing (~0.11 ms), zero-shot local inference (~0.15 ms), and deterministic stitching (~0.009 ms). The online latency is therefore below the effective acquisition interval, enabling a solve-while-scanning workflow in which phase contrast can be updated during data collection. Importantly, this streaming mode does not require explicit probe input or online iterative refinement during inference. The probe and overlap information are instead inherited through the ePIE-derived teacher labels used during training. For offline processing of large 4D-STEM datasets, optimized batched inference further reduces the per-frame inference latency to ~0.049 ms (Fig. 1e). Benchmarking across scan grids from 50 × 50 to 700 × 700 positions shows stable high-throughput processing at ~20,000 probe positions per second. Relative to GPU-accelerated ePIE benchmarked at 15 iterations, totaling ~53.6 ms per frame based on ~3.57 ms per frame per iteration, this corresponds to an approximately 1,090-fold speedup. This speed gain is not obtained by learning an unconstrained global image-to-image mapping. Instead, the network is trained on local ePIE-derived reference phase patches within the probe coherent core, so the inference target remains tied to a physics-constrained ptychographic teacher. Once trained on a single experimental AuPd dataset, the same model is used for all subsequent zero-shot tests without fine-tuning, retraining, or explicit probe recalibration.

**Spatial hold-out validation on AuPd under matched imaging conditions**

We next established the baseline fidelity of the trained model under matched imaging conditions using a strict spatial hold-out protocol on the AuPd dataset (Fig. 2). The model was trained on experimental AuPd acquired at spot size 9 and df = 30 nm, while the bottom-right 200 × 200 quadrant of the original 400 × 400 scan was fully excluded from training and reserved for testing (Fig. 2a,b). A total of 15,000 diffraction frames were randomly sampled from the remaining training pool. This spatial split is more stringent than a random frame-wise partition because it prevents direct leakage from neighboring scan positions in the held-out region.

In the held-out AuPd region, the AI inference closely reproduces the ePIE reference phase map under the same imaging condition (Fig. 2b–d). At the full-field scale, the inferred phase image preserves the mesoscale contrast variations and local lattice ordering observed in the ptychographic reference. In the representative local region, the AI inference maintains atomic-scale lattice contrast consistent with ePIE, while HAADF is used only as a qualitative structural reference and not as the target for phase evaluation (Fig. 2d).

Reciprocal-space analysis further supports this agreement. The FFT magnitude maps of the representative region show closely matched peak positions and comparable reciprocal-space broadening between AI inference and the ePIE reference, indicating preservation of the dominant lattice periodicity and grain-scale structural information (Fig. 2d). Patch-level quantitative analysis shows tightly distributed PSNR and MS-SSIM values in the held-out region (Fig. 2c). FRC analysis between AI inference and the ePIE reference yields a cutoff of 0.68 Å for the representative region (Fig. 2e), confirming that the learned local estimator preserves high-frequency phase information on a spatially unseen region of the same experimental dataset. Additional spatial-partition and preprocessing ablations further confirm that this matched-condition performance is not tied to one privileged training quadrant or to rotational preprocessing alone (Supplementary Fig. 3).

**Cross-material zero-shot transfer from AuPd to $WS_2$**

To test whether the learned local prior is limited to AuPd or can transfer across material systems, we applied the frozen AuPd-trained model directly to experimental $WS_2$ data without fine-tuning (Fig. 3). This transfer represents a clear domain shift from a polycrystalline metallic AuPd sample to a layered hexagonal transition metal dichalcogenide with different symmetry, scattering behavior, and local structural motifs (Fig. 3a). Despite this shift, the model recovers the characteristic $WS_2$ lattice directly from CBED measurements and produces phase contrast consistent with the ePIE reference (Fig. 3b,c).

In the TOP and BOTTOM regions marked in Fig. 3b, the AI inference preserves the dominant lattice periodicity observed in the ePIE reference (Fig. 3c). The FFT magnitude maps show matched Bragg-peak positions, with $q_{max} \approx 0.821$ Å$^{-1}$ in both representative regions, corresponding to a real-space periodicity of approximately 1.22 Å. The reciprocal-space peak widths in the AI result remain close to those of the ePIE reference, indicating that local periodicity and grain-scale information are retained after zero-shot transfer. FRC analysis further confirms high-frequency agreement with the ptychographic reference, with cutoffs of 0.67 Å and 0.72 Å for the TOP and BOTTOM regions, respectively (Fig. 3d). Quantitative evaluation over 20 randomly sampled regions showed closely overlapping PSNR and MS-SSIM distributions between AI inference and the ePIE reference, confirming stable cross-material fidelity (Fig. 3e).

Beyond periodic lattice regions, we evaluated the same frozen model on a defect-containing $WS_2$ region with a large hole-like feature (Fig. 3f–h). The AI inference follows the vacancy boundary and preserves phase contrast consistent with the ePIE reference across the crystalline-to-void transition (Fig. 3g). Fourier analysis shows that the dominant lattice frequency is retained, with $q_{max} \approx 0.628$ Å$^{-1}$, corresponding to a real-space periodicity of approximately 1.59 Å. Although the AI Fourier peak is slightly broader than that of the ePIE reference in this defect-containing region, the dominant periodicity remains visible and the reconstruction does not collapse into a purely periodic texture. The high-frequency branch ratio map further shows adaptive spectral modulation near the defect boundary, with reduced high-frequency emphasis inside the hole region where lattice information is absent (Fig. 3g). Horizontal and vertical line profiles across the marked defect region show that the AI inference follows the main phase variation of the ePIE reference while producing a cleaner response inside the void (Fig. 3h). These results indicate that the AuPd-trained model does not simply memorize a material-specific appearance, but learns local scattering regularities that transfer to both periodic and structurally heterogeneous $WS_2$ regions.

**Zero-shot phase inference under continuous defocus-induced transfer-function modulation**

We then tested the stability of the learned local prior under continuous transfer-function changes using a defocus series of high-entropy alloy nanoparticles (HEA-NP) (Fig. 4). The model was trained only on AuPd acquired at spot size 9 and df = 30 nm, then frozen and applied to HEA-NP datasets acquired at the same spot size but across df = 0–40 nm without fine-tuning or explicit probe input (Fig. 4a). Varying defocus changes the probe phase curvature and reshapes the intensity distribution within the diffraction disks, providing a direct test of robustness to probe-transfer modulation.

The largest defocus shift from the training condition occurs at df = 0 nm. Under this condition, the AI inference still reproduces the main atomic lattice contrast observed in the ePIE reference (Fig. 4b). Although pixel-wise agreement is lower than under the matched df = 30 nm condition, the local FFT magnitude maps retain the principal reciprocal-space peaks, indicating preservation of high-frequency structural information. FRC analysis at df = 0 nm yields a cutoff of 0.77 Å relative to the ePIE reference (Fig. 4d), supporting structural consistency under the strongest tested defocus shift.

Across the full df = 0–40 nm series, local real-space and reciprocal-space comparisons show that the AI inference preserves the main lattice contrast and peak structure across representative defocus levels (Fig. 4c). Quantitative evaluation shows the expected trend: PSNR and MS-SSIM reach their highest values near the training condition of df = 30 nm, while structural agreement remains stable across the tested range (Fig. 4e). Thus, the learned local estimator has tolerance to continuous transfer-function modulation within the tested defocus range. Rather than acting as a fully general defocus corrector, the model should be understood as learning a local ptychographic-reference prior that remains usable under moderate defocus-induced probe changes.

**Cross-instrument zero-shot transfer and sparse-sampling robustness**

Finally, we evaluated cross-instrument generalization and sparse-sampling robustness using hBN and step-size variation experiments (Fig. 5). The model trained on AuPd acquired at 200 kV and spot size 9 was applied directly to an hBN dataset acquired on a distinct 300 kV microscope with a reported convergence semi-angle of 30 mrad, without fine-tuning or explicit probe input during inference (Fig. 5a). This test introduces changes in material composition, atomic number, microscope voltage, convergence angle, probe condition, detector configuration, and acquisition statistics. Despite these differences, the AI inference reconstructs the hBN lattice with phase contrast closely comparable to the ePIE reference (Fig. 5b). We further compared the proposed method with the DPI baseline under the same hBN evaluation setting (Fig. 5b,c). The DPI reconstruction shows stronger smoothing and weaker atomic contrast under these experimental domain shifts, while our ptychography-supervised local inference retains clearer lattice features. In reciprocal tests on our experimental AuPd and $WS_2$ datasets, the DPI baseline also failed to achieve the same sub-ångström-level agreement with the ePIE reference (Supplementary Fig. 1). These comparisons support the interpretation that probe-specific or simulation-trained supervised mappings can be less stable when transferred to experimental data acquired under different probe and instrument conditions.

To examine the role of physical overlap during deployment, we evaluated performance under increasing scan step sizes from 0.25 Å to 1.0 Å (Fig. 5c,d). As the step size increases, the physical overlap between neighboring probe positions decreases, making full-field assembly more difficult. Our method retains clearer local lattice contrast than DPI across the tested sampling densities, and PSNR/MS-SSIM distributions show consistently higher agreement with the ePIE reference (Fig. 5d). However, reduced overlap progressively limits field continuity and stitching stability. This behavior is consistent with the proposed workflow: local phase prediction is performed without solving an online global ptychographic inverse problem, but full-field phase imaging still requires sufficient geometric sampling and relative scan-coordinate consistency. Thus, the degradation at sparse sampling should be understood mainly as a loss of geometric continuity during assembly, rather than a complete failure of local phase inference. These results define both the practical strength and the boundary of the method: it can support real-time and cross-condition phase imaging, but it still relies on adequate spatial sampling to form a continuous object field.

## Discussion

We have presented a ptychography-supervised local phase imaging framework that converts high-overlap electron ptychographic reconstruction into fast, probe-prior-free inference. The central point of this work is not that the physical constraints of ptychography are removed, but that their role is shifted within the imaging workflow. In conventional iterative reconstruction, spatial overlap, probe relaxation, scan-coordinate consistency, and global object continuity are enforced online for each newly acquired 4D-STEM dataset. In our framework, these constraints are first used by iterative ePIE reconstruction to generate physics-constrained teacher phase maps. The neural model then learns the local regularities contained in these teacher reconstructions and amortizes them into a compact phase estimator. During deployment, the model predicts local ePIE-consistent phase patches from individual diffraction measurements within the probe coherent core, while overlap is used mainly to assemble these local predictions into a continuous field. Thus, physical redundancy remains essential, but it is used differently: it is embedded into the teacher reconstruction and learned representation, rather than repeatedly solved through online global optimization.

This formulation also clarifies the relation between our method and traditional electron ptychography. Classical constraints such as the Faulkner-Rodenburg overlap condition[8] address the non-uniqueness of diffraction phase retrieval by enforcing consistency among many overlapping measurements. Our method does not claim that this ambiguity is generally resolved from a single diffraction frame in a purely mathematical sense. Instead, we show that, within the experimentally sampled distribution and the coherent core of the probe, local phase patches reconstructed by high-overlap ptychography occupy a restricted statistical space that can be learned from data. The network therefore performs statistically regularized local inference with respect to a ptychographic teacher, rather than unconstrained single-frame phase retrieval. The output is best understood as an ePIE-consistent local phase estimate, whose physical basis comes from the ptychographic reconstruction used for supervision.

The zero-shot transfer results suggest that this learned local prior is not limited to memorizing one material texture or one scan region. A model trained only on experimental AuPd data generalizes to $WS_2$, defocus-varied high-entropy alloy nanoparticles, and cross-instrument hBN measurements. This observation supports a finite-manifold view of local phase imaging. When the inference window is limited to the scale of the probe coherent core, the model does not need to represent arbitrary extended object potentials. Instead, it learns a restricted set of local scattering configurations that are repeatedly sampled across thin crystalline and nanostructured specimens. In this sense, polycrystalline AuPd acts as a structurally rich experimental source: it contains many local orientations, grain boundaries, contrast variations, and short-range structural motifs. These variations provide broad coverage of the ptychographic-reference phase manifold, allowing the network to learn transferable local scattering regularities rather than a single material-specific template.

Supplementary Fig. 5 further supports this interpretation by showing that zero-shot performance depends strongly on the statistical richness of the training source. Models trained on different material systems, including $WS_2$, HEA-NP, and hBN, were evaluated on the same held-out AuPd region. The AuPd-trained model achieved the highest fidelity, while the HEA-NP-trained model retained clear lattice periodicity and reciprocal-space peak structure. In contrast, models trained on highly ordered $WS_2$ and hBN data showed weaker lattice contrast and broader Fourier features. This trend does not simply follow chemical identity. Rather, it suggests that training data with richer local structural diversity provide better coverage of the local scattering manifold. This view is also consistent with the training-size analysis in Supplementary Fig. 4, where zero-shot performance improves with increasing AuPd patches but approaches saturation at approximately $1.5 \times 10^4$ experimental patches. Beyond this regime, additional data do not consistently improve transfer and may introduce mild source-specific specialization.

The observed generalization is not automatic; it depends on both the teacher reconstruction and the physical biases built into the model. Electron ptychography differs from many X-ray ptychography settings because the electron probe can be affected by source coherence, aberration residuals, defocus, and other microscope-dependent factors[26-28]. A direct network mapping may therefore learn coupled probe-object features rather than a transferable object-phase representation[16]. Our architecture was designed to reduce this risk. The low-frequency branch captures the bright-field envelope and slowly varying Ronchigram-like modulations, while the high-frequency branch focuses on interference fringes and lattice-related information needed for atomic-resolution recovery. Polar positional encoding and rotational augmentation encourage the model to treat rotated diffraction features consistently and reduce sensitivity to anisotropic probe artifacts. The domain-attention fusion module adapts the balance between low-frequency envelope information and high-frequency fringe information according to the local diffraction signal. Together with rapid convergence, which can act as implicit regularization[29], these choices help the model learn transferable local scattering features before memorizing source-specific noise.

This operational shift changes the efficiency regime of electron phase imaging. As summarized in Supplementary Table 1, the proposed framework differs from probe-specific supervised baselines in training source, probe requirement, ground-truth definition, model size, training time, global reconstruction strategy, and cross-domain behavior. The main difference is not only speed, but where the hard computation occurs. Traditional ePIE solves a coupled inverse problem for every new scan, so reconstruction time grows with the number of probe positions and the number of iterations required for convergence. In contrast, our model performs the expensive ptychographic optimization during teacher generation and training, then uses a single forward pass for local prediction and deterministic overlap stitching for field assembly. This amortized design makes the online latency compatible with detector acquisition, enabling a solve-while-scanning workflow. The broader value is that electron ptychography can move from offline post-processing toward live phase observation, where phase contrast can guide field-of-view selection, dose control, drift assessment, and high-throughput materials screening.

The sparse-scanning experiment clarifies the remaining role of overlap. At inference time, local phase prediction does not depend on absolute scan history or on solving a global optimization problem. However, a continuous full-field image still requires relative spatial placement and sufficient overlap for stable assembly. When the step size increases and overlap decreases, the main failure mode is loss of geometric continuity during stitching, rather than a complete collapse of local phase prediction. This behavior indicates that overlap has not disappeared from the imaging process. Its online role is shifted from driving iterative phase retrieval to supporting field assembly and boundary consistency. Supplementary Fig. 3 further shows that the in-distribution AuPd result is not caused by leakage from one privileged scan region. Models trained from different spatial quadrants recover similar local lattice morphology and comparable reciprocal-space peak structure on the same held-out region, indicating that the learned mapping is spatially stationary within the AuPd scan.

Several limitations should be made explicit. First, the method is supervised by ePIE-derived reference phases. These references are physics-constrained, but they are not independent experimental ground truth. The network may therefore inherit teacher bias, including probe assumptions, phase wrapping behavior, position correction, and residual artifacts in the iterative solution. Quantitative metrics such as PSNR, MS-SSIM, and FRC should be interpreted as agreement with the ePIE reference, not as direct measurement of an unknown true phase. Second, the current model is validated mainly in thin-sample and weak-to-moderate scattering regimes. Under these conditions, local diffraction features remain sufficiently related to projected phase contrast for the learned mapping to transfer across materials and probe variations. In regimes dominated by strong multiple scattering, such as thicker zone-axis crystals, the projection-like local phase model may no longer be adequate. Extending the approach to such cases will likely require teacher reconstructions and neural models that include multi-slice physics or depth-dependent scattering representations. Although the model shows tolerance to probe, defocus, and instrument changes within the tested range, it is not fully probe-independent in an unrestricted sense. The in-focus spot-size-7 AuPd experiment in Supplementary Fig. 2 illustrates this boundary. Under this larger probe-domain shift, the AI reconstruction still preserves the dominant lattice periodicity and reaches an FRC cutoff of 0.78 Å relative to the ePIE reference, but the local phase morphology becomes less stable than under matched training conditions. Some discrepancies are concentrated near sharp phase transitions and wrapped phase boundaries, where the ePIE reference itself can show split minima or local fluctuations at atomic sites. This result shows that the learned prior has a finite domain of validity and should not be interpreted as universally probe-agnostic.

The low-dose analysis in Supplementary Fig. 6 defines another boundary. The AuPd-trained model remains structurally stable from the measured $WS_2$ input down to $\alpha = 10^{-5}$, but degrades sharply at $\alpha = 10^{-6}$, where the high-frequency diffraction information is dominated by counting noise. This transition reflects a physical information limit rather than an arbitrary neural-network instability.

Together, these results position the proposed method as an experimental-data-driven, ptychography-supervised route to real-time electron phase imaging. Its main contribution is not a replacement of ptychography by a black-box network, but an amortization of ptychographic redundancy into a deployable local inference model. For electron microscopy, the immediate value lies in live phase feedback, larger practical fields of view, and faster screening under realistic probe and drift variations. More generally, the same strategy may be useful in other coherent imaging settings where a rigorous but slow reconstruction method can provide teacher reconstructions, and where local measurement-image relationships are sufficiently constrained by experimental structure. Future work will extend this framework to stronger dynamical scattering, improved uncertainty estimation, adaptive scan control, and teacher models that include multi-slice physics, with the long-term goal of making high-resolution phase imaging available during data acquisition rather than only after offline reconstruction[30].

## Methods

### Sample preparation

Four representative material systems were used to evaluate the proposed ptychography-supervised local phase imaging framework across structural, compositional, and instrumental domains. The AuPd thin film, composed of alloy nanoparticles, was deposited onto a carbon replica substrate using a gold-palladium alloy target in a DC sputter coater. AuPd was selected as the training source because its polycrystalline nanoparticle structure provides a broad distribution of local orientations, grain boundaries, and short-range phase variations, which is useful for sampling diverse local scattering configurations. Tungsten disulfide ($WS_2$) specimens were prepared by mechanical exfoliation from commercial bulk crystals (HQ Graphene) using polydimethylsiloxane (PDMS) sheets. The exfoliated flakes were transferred onto holey silicon nitride TEM grids (Norcada) using a modified viscoelastic stamp process, producing suspended atomically thin transition metal dichalcogenide regions suitable for high-resolution scanning transmission electron microscopy (STEM). PtMoCu high-entropy alloy nanoparticles were synthesized at high temperature, rinsed, and immersed in deionized water to remove residual reactants. The resulting nanoparticles were deposited onto ultrathin carbon film transmission electron microscopy (TEM) grids for imaging. For cross-instrument evaluation, an open-access monolayer hexagonal boron nitride (hBN) dataset was included. This dataset represents a low-Z crystalline specimen acquired on a distinct 300 kV electron microscope with a reported convergence semi-angle of 30 mrad. Together, these samples cover different crystal symmetries, scattering strengths, probe conditions, and instrument configurations, providing a basis for testing zero-shot generalization beyond the AuPd training distribution.

**Data acquisition and experimental datasets**

All in-house datasets were experimentally acquired to preserve realistic 4D-STEM signal statistics, detector response, probe variations, and microscope-dependent effects. Four material systems were used: AuPd for supervised training, and $WS_2$, high-entropy alloy nanoparticles (HEA-NP), and hBN for zero-shot evaluation. The AuPd training dataset was acquired using a Thermo Fisher Titan G2 80-200 ChemiSTEM operated at 200 kV with a Gatan STELA detector. The acquisition used spot size 9, defocus df = 30 nm, 512 × 512 pixel diffraction frames, a 200 μs exposure per probe position, a 0.5 Å real-space step size, and a 400 × 400 scan grid. From approximately 160,000 raw diffraction frames, 15,000 frames were randomly selected from the training region to construct paired diffraction-phase samples. Random rotational augmentation produced 75,000 effective training samples. The $WS_2$, HEA-NP, and hBN datasets were acquired or obtained under distinct material, voltage, defocus, probe, and detector conditions and were used without fine-tuning or retraining. Variations in spot size and defocus change the illumination wavefront by modifying phase curvature, coherent-core size, and the effective overlap relation between adjacent probe positions. The hBN data were obtained from the open-access dataset reported by Chang et al.[16]. Detailed acquisition parameters, detector formats, readout binning, and instrument settings are summarized in Supplementary Tables 2-3 and Supplementary Figs. 8-9.

**Data preprocessing and patch construction**

The neural model was trained using ePIE-derived reference phase patches rather than independent experimental ground truth. For each selected probe position, the corresponding local phase patch was extracted from a full-field coordinate-corrected ePIE reconstruction and spatially registered to the raw convergent-beam electron diffraction (CBED) frame. The raw diffraction frames were binned from 512 × 512 to 64 × 64 pixels before patch construction to improve the signal-to-noise ratio and reduce computational cost. To preserve the geometric relation between reciprocal-space diffraction and real-space phase, randomized rotations were applied simultaneously to the paired diffraction and reference phase tensors before final cropping.

After augmentation, each CBED pattern was decomposed into two input channels using a dynamic 93rd-percentile intensity threshold. The high-intensity channel captures the bright-field disk and Ronchigram-like envelope, which contain low-frequency shadowing and probe-specimen interaction information. The low-intensity channel retains weaker scattering fringes and higher-frequency diffraction information that are important for atomic-resolution phase contrast. The decomposed diffraction inputs were center-cropped to 32 × 32 pixels and normalized by channel-wise Z-score standardization. The corresponding ePIE-derived reference phase target was cropped to a central 16 × 16 pixel window. This output size was selected to match the physical extent of the probe coherent core, as discussed in Supplementary Text 1. By restricting the prediction target to the local interaction region, the model is trained to infer local ePIE-consistent phase patches rather than an extended global object field.

**Model architecture and design rationale**

The Multi-domain Representation and Domain-Attention Model (MRDM) is a dual-encoder network designed for ptychography-supervised local phase inference (Supplementary Fig. 7). The network takes a 32 × 32 dual-channel diffraction patch as input and predicts a local phase patch associated with the probe coherent core. The architecture was designed to separate low-frequency envelope features from high-frequency interference information, reducing the risk that the model learns a probe-specific intensity pattern instead of transferable local scattering regularities. The low-frequency (LF) branch uses a convolutional encoder with atrous spatial pyramid pooling (ASPP)[31] and squeeze-and-excitation (SE)[32] modules. This branch captures the bright-field envelope, Ronchigram-like contrast, and slowly varying local modulations, which help stabilize predictions in low-contrast or weakly scattering regions. The high-frequency (HF) branch uses a Transformer encoder[33],[34] with polar positional encoding ($r$, $\theta$, $\sin\theta$, $\cos\theta$). This branch focuses on weak interference fringes and lattice-related high-frequency diffraction features. A radial gate with r_bias ≈ 0.55 and a band-pass region $r \in [0.5, 1.0]$ suppresses noise-dominated components while retaining structurally relevant scattering information. The LF and HF embeddings are fused using a domain-attention module. The adaptive weights are computed as

$$[\alpha_{LF}, \alpha_{HF}] = softmax(\frac{Wconcat(s_{LF}, s_{HF})}{\tau}),\ \alpha_{LF} + \alpha_{HF} = 1,\ \tau = 0.6.$$

The fused representation is given by $f_{fused} = \alpha_{LF}c_{LF} + \alpha_{HF}c_{HF}$. This design provides interpretable attention weights that describe the balance between envelope-dominated and fringe-dominated information in each local diffraction pattern. Rotational augmentation and polar encoding encourage consistent treatment of rotated diffraction features and help reduce sensitivity to anisotropic probe artifacts such as astigmatism. A lightweight convolutional decoder maps the fused representation to the final local phase prediction without upsampling, preserving the local spatial scale of the coherent-core phase patch. The architecture therefore acts as a physically motivated local estimator that learns the statistical relation between experimental diffraction features and ePIE-derived phase patches.

**Uncertainty quantification and interpretability**

Predictive uncertainty was estimated using Monte Carlo dropout[35] during inference. For each input diffraction patch, ten stochastic forward passes (N = 10) were performed, and the pixel-wise predictive variance Var(y) was computed. The resulting uncertainty maps were used to identify regions where the model response was less stable, such as interference boundaries, defect edges, and other locally ambiguous structures. In the $WS_2$ defect-region analysis, elevated uncertainty was spatially correlated with domain boundaries and non-periodic transition regions, suggesting that the uncertainty reflects physical ambiguity in the diffraction signal rather than only numerical noise.

**Training and optimization strategy**

All main experimental results were obtained using a single model trained from scratch on one experimentally acquired AuPd 4D-STEM dataset with spot size 9, df = 30 nm, and 200 kV acceleration voltage. No synthetic diffraction data, explicit probe priors, or multi-dataset training were used. This setting was chosen to test whether a structurally rich experimental source can provide sufficient local scattering diversity for zero-shot transfer. The model was trained using the Adam optimizer with learning rate $1 \times 10^{-4}$, $\beta_1 = 0.9$, and $\beta_2 = 0.999$. The primary loss was the mean absolute error (MAE) between the predicted local phase patch and the ePIE-derived reference patch. To further encode rotational consistency, a rotation-consistency regularizer was included:

$$\mathcal{L}_{\mathrm{rot}} = \| R_{-\theta}\left( y_{R_\theta} \right) - y \|_1,$$

and the total loss was

$$\mathcal{L}_{\mathrm{total}} = \mathcal{L}_{\mathrm{MAE}} + 0.1\mathcal{L}_{\mathrm{rot}}.$$

Each mini-batch contained 32 patches. Training was performed for no more than 3 epochs with a learning-rate scheduler using a reduction factor of 0.2. The same trained model was used for all reported tests, including spatially held-out AuPd, $WS_2$, HEA-NP defocus series, and hBN cross-instrument evaluation. No fine-tuning, retraining, or dataset-specific adaptation was applied during testing.

**Global iterative-optimization-free phase stitching**

To obtain full-field phase images, the locally predicted phase patches were assembled using a deterministic stitching protocol. Unlike ePIE, which enforces global consistency through iterative optimization, this step treats full-field assembly as a linear accumulation process. For adjacent predicted patches, the constant phase offset was estimated by calculating the mean phase difference within the physical overlap region. After offset alignment, patches were merged using a weighted overlap-add operation. This procedure performs only addition, averaging, and boundary smoothing, and does not use gradient-based optimization or iterative refinement. The computational cost therefore scales linearly with the number of scan positions, O(N), which supports high-throughput and streaming reconstruction. In this workflow, overlap remains important for field continuity and boundary consistency, but it is no longer used to solve an online global inverse problem during deployment.

**Reference reconstruction and teacher-label generation**

The reference phase reconstructions used for training and benchmarking were generated using the extended ptychographic iterative engine (ePIE)[9], accelerated by GPU implementation. Before reconstruction, the diffraction patterns were binned from 512 × 512 to 64 × 64 pixels to improve the signal-to-noise ratio. Two assumed incoherent probe modes were used in the mixed-state reconstruction, following the formulation described by Chen et al.[27]. The object and probe updates can be written as

$$\alpha O_r' = O_r + \frac{\underset{k}{max} P(k)_r^2 P(k)_r^* \times (\psi(k)_r - P(k)_r O_r)}{\underset{k}{max} P(k)_r^2}$$

$$P(k)_r' = P(k)_r + \beta \frac{max|O_r|^2 O_r^* \times (\psi(k)_r - P(k)_r O_r)}{max|O_r|^2}$$

Here, $O_r$ denotes the object function, $P(k)_r$ denotes the k-th probe mode, $\psi(k)_r$ denotes the exit wave in real space, and the superscript * indicates complex conjugation. The constants α and β regulate the object and probe update strengths and were chosen according to experimental reconstruction stability. To generate coordinate-consistent reference phase maps suitable for local neural training, we used a probe-centered reconstruction procedure consisting of 15 joint ePIE iterations, probe recentering, 5 fixed-probe iterations, and a final 15 unconstrained iterations, for 35 total iterations. This protocol suppresses lateral probe drift and aligns the illumination basis to a common center. The resulting coordinate-corrected phase maps provide locally comparable ePIE-derived reference patches across the scan. These patches are used as teacher labels for the neural model and as the reference for quantitative benchmarking. They should be interpreted as ptychographic-reference phases rather than independent experimental ground truth.

**Evaluation and metrics computation**

All quantitative evaluations were performed on stitched full-field phase reconstructions. Before metric computation, the AI-stitched reconstruction and the ePIE reference were aligned using sub-pixel phase correlation to correct residual translational offsets. After alignment, both images were normalized to the range [0,1]. Pixel-wise metrics, including mean squared error (MSE) and peak signal-to-noise ratio (PSNR), were computed as

$$\mathrm{MSE} = \frac{1}{N}\sum_{i}(A_{\mathrm{AI},i} - A_{\mathrm{GT},i})^2, \mathrm{PSNR} = 10\log_{10}(\frac{1}{\mathrm{MSE}}).$$

Structural similarity index (SSIM) and multi-scale SSIM (MS-SSIM)[36] were computed within corresponding regions of interest. MS-SSIM was evaluated using an 11 × 11 Gaussian window to capture structural similarity across spatial frequencies. Fourier-domain analysis was performed using FFT magnitude maps, Bragg-peak positions, and Fourier ring correlation (FRC). FRC was computed only between the AI reconstruction and the stitched ePIE reference. HAADF images were used solely as qualitative real-space references and were not used as ground truth for phase metrics.

**Computational resources and performance characterization**

Reference ePIE reconstructions were performed on a workstation equipped with an AMD Ryzen 9 5950X CPU with 16 cores, 64 GB RAM, and an NVIDIA RTX A6000 GPU with 48 GB memory. For a standard 400 × 400 scan grid, one ePIE iteration required 571.47 ± 3.51 s, measured as the trimmed mean of 15 independent runs. This corresponds to approximately 3.57 ms per frame per iteration. For performance benchmarking, 15 ePIE iterations were used, resulting in an effective processing latency of approximately 53.6 ms per frame and about 2.4 hours for the full field. Although the full teacher reconstruction used 35 iterations for label generation, the 15-iteration setting was used as a practical speed baseline for comparison with online inference.

The proposed neural model was evaluated on a lightweight cloud instance equipped with an Intel Xeon CPU with 2 virtual cores and an NVIDIA Tesla T4 GPU with 15 GB memory. This lower-cost hardware was used to test deployment feasibility under resource-constrained conditions. Performance values were obtained from 15 independent trials, reporting trimmed means after removing the maximum and minimum values. Preprocessing and stitching latencies were measured over the full 400 × 400 grid. Inference latency was measured using 1,000 consecutive batches per trial to reduce thermal and scheduler fluctuations. For the online streaming regime, the timing protocol included the complete data round trip, including host-to-device transfer, device-to-host synchronization, PCIe bandwidth effects, and CPU blocking overhead. A batch size of 16 was selected because it balanced bus-cost amortization and buffering delay. Under this setting, preprocessing, inference, and stitching gave a total latency of approximately 0.27 ms per frame, which approaches the 0.22 ms detector dwell time and remains below the effective 0.50 ms sampling interval including I/O. Full batch-size scaling is reported in Supplementary Table 4. To estimate the intrinsic algorithmic upper bound, offline throughput was measured using XLA compilation and pre-allocated GPU memory with a saturated batch size of 1,024. This reduced inference latency to approximately 0.049 ms per frame, corresponding to about 20,000 probe positions per second. These benchmarks show that the proposed method can match practical detector acquisition rates in the streaming setting while also supporting high-throughput offline reconstruction for large 4D-STEM scans.

**Acknowledgements**
The authors would like to thank National Tsing Hua University and the National Center for Instrumentation Research for providing access to research facilities and technical support.

**Author contributions**
C.-C. Chen and F. Shu conceived the research. H. Yue and F. Shu carried out the machine learning and model development; H. Yue implemented the deep learning framework and performed the network training, validation, and visualized the results. C.-C. Chen and C.-N. Hsiao prepared the samples and conducted the data acquisition experiments. C.-C. Chen analyzed the data and performed the reference reconstructions. J. Cheng contributed to the network design and generalization validation. Y. Liu and X.Z. Liao provided guidance on the materials science implications and helped on model interpretability and limitations. H. Yue, C.-C. Chen, and F. Shu wrote the manuscript; All authors reviewed the manuscript.

**Competing interests**
The authors declare no competing interests.

**Funding**
This work was supported by the Angstrom Semiconductor Initiative Project of the National Science and Technology Council (NSTC), Taiwan, under Grant No. NSTC114-2119-M-007-016-MBK.

**Data availability**
The datasets used and/or analysed during the current study are available from the corresponding author on reasonable request.

**Code availability**
Source code, pretrained weights, and minimal preprocessing scripts will be released upon publication to ensure full reproducibility.

**References**

1. Ophus, C. Four-dimensional scanning transmission electron microscopy (4D-STEM): From scanning nanodiffraction to ptychography and beyond. Microscopy and Microanalysis 25, 563-582 (2019).
2. Tate, M. W. et al. High dynamic range pixel array detector for scanning transmission electron microscopy. Microscopy and Microanalysis 22, 237-249 (2016).

3. Nellist, P. D., McCallum, B. C. & Rodenburg, J. M. Resolution beyond the 'information limit' in transmission electron microscopy. Nature 374, 630-632 (1995).
4. Jiang, Y. et al. Electron ptychography of 2D materials to deep sub-ångström resolution. Nature 559, 343-349 (2018).
5. Chen, Z. et al. Electron ptychography achieves atomic-resolution limits set by lattice vibrations. Science 372, 826-831 (2021).
6. Miao, J. Computational microscopy with coherent diffractive imaging and ptychography. Nature 637, 281-295 (2025).
7. Bunk, O. et al. Influence of the overlap parameter on the convergence of the ptychographical iterative engine. Ultramicroscopy 108, 481-487 (2008).
8. Rodenburg, J. M. & Faulkner, H. M. A phase retrieval algorithm for shifting illumination. Applied Physics Letters 85, 4795-4797 (2004).
9. Maiden, A. M. & Rodenburg, J. M. An improved ptychographical phase retrieval algorithm for diffractive imaging. Ultramicroscopy 109, 1256-1262 (2009).
10. Fienup, J. R. Phase retrieval algorithms: a comparison. Applied Optics 21, 2758-2769 (1982).
11. Cao, M. C., Chen, Z., Jiang, Y. & Han, Y. Automatic parameter selection for electron ptychography via Bayesian optimization. Scientific Reports 12, 12284 (2022).
12. Wang, K. et al. On the use of deep learning for phase recovery. Light: Science & Applications 13, 4 (2024).
13. Babu, A. V. et al. Deep learning at the edge enables real-time streaming ptychographic imaging. Nature Communications 14, 7059 (2023).

14. Guan, Z., Tsai, E. H., Huang, X., Yager, K. G. & Qin, H. Ptychonet: Fast and High Quality Phase Retrieval for Ptychography. https://www.osti.gov/biblio/1599580 (2019).

15. Cherukara, M. J. et al. AI-enabled high-resolution scanning coherent diffraction imaging. Applied Physics Letters 117, (2020).

16. Chang, D. J. et al. Deep-Learning Electron Diffractive Imaging. Physical Review Letters 130, 016101 (2023).

17. Yue, H. et al. A Physics-Inspired Deep Learning Framework With Polar Coordinate Attention for Ptychographic Imaging. IEEE Transactions on Computational Imaging (2025).

18. Antun, V., Renna, F., Poon, C., Adcock, B. & Hansen, A. C. On instabilities of deep learning in image reconstruction and the potential costs of AI. Proc. Natl. Acad. Sci. U.S.A. 117, 30088-30095 (2020).

19. Belthangady, C. & Royer, L. A. Applications, promises, and pitfalls of deep learning for fluorescence image reconstruction. Nature Methods 16, 1215-1225 (2019).

20. Pan, X. *et al.* An efficient ptychography reconstruction strategy through fine-tuning of large pre-trained deep learning model. *Iscience* **26**, (2023).

21. Ongie, G. *et al.* Deep learning techniques for inverse problems in imaging. *IEEE Journal on Selected Areas in Information Theory* **1**, 39–56 (2020).

22. Du, M., Huang, X. & Jacobsen, C. Using a modified double deep image prior for crosstalk mitigation in multislice ptychography. *Journal of Synchrotron Radiation* **28**, 1137–1145 (2021).

23. Williams, G. J. *et al.* Fresnel coherent diffractive imaging. *Physical Review Letters* **97**, 025506 (2006).

24. Abbey, B. *et al.* Keyhole coherent diffractive imaging. *Nature Physics* **4**, 394–398 (2008).

25. Brown, M. & Lowe, D. G. Automatic Panoramic Image Stitching using Invariant Features. *Int J Comput Vision* **74**, 59–73 (2007).
26. Maiden, A. M., Humphry, M. J., Sarahan, M. C., Kraus, B. & Rodenburg, J. M. An annealing algorithm to correct positioning errors in ptychography. *Ultramicroscopy* **120**, 64–72 (2012).
27. Chen, Z. *et al.* Mixed-state electron ptychography enables sub-angstrom resolution imaging with picometer precision at low dose. *Nature Communications* **11**, 2994 (2020).
28. Odstrcil, M. *et al.* Ptychographic coherent diffractive imaging with orthogonal probe relaxation. *Optics Express* **24**, 8360–8369 (2016).
29. Ulyanov, D., Vedaldi, A. & Lempitsky, V. Deep image prior. in *Proceedings of the IEEE conference on computer vision and pattern recognition* 9446–9454 (2018).
30. Zheng, G., Horstmeyer, R. & Yang, C. Wide-field, high-resolution Fourier ptychographic microscopy. *Nature Photonics* **7**, 739–745 (2013).
31. Chen, L.-C., Papandreou, G., Kokkinos, I., Murphy, K. & Yuille, A. L. Deeplab: Semantic image segmentation with deep convolutional nets, atrous convolution, and fully connected crfs. *IEEE Transactions on Pattern analysis and Machine Intelligence* **40**, 834–848 (2017).
32. Hu, J., Shen, L. & Sun, G. Squeeze-and-excitation networks. in *Proceedings of the IEEE conference on computer vision and pattern recognition* 7132–7141 (2018).
33. Vaswani, A. *et al.* Attention is all you need. *Advances in neural information processing systems* **30**, (2017).
34. Dosovitskiy, A. *et al.* An Image Is Worth 16x16 Words: Transformers for Image Recognition at Scale. (2021).

35. Gal, Y. & Ghahramani, Z. Dropout as a bayesian approximation: Representing model uncertainty in deep learning. in *international conference on machine learning* 1050–1059 (PMLR, 2016).

36. Wang, Z., Simoncelli, E. P. & Bovik, A. C. Multiscale structural similarity for image quality assessment. in *The thrity-seventh asilomar conference on signals, systems & computers, 2003* vol. 2 1398–1402 (Ieee, 2003).

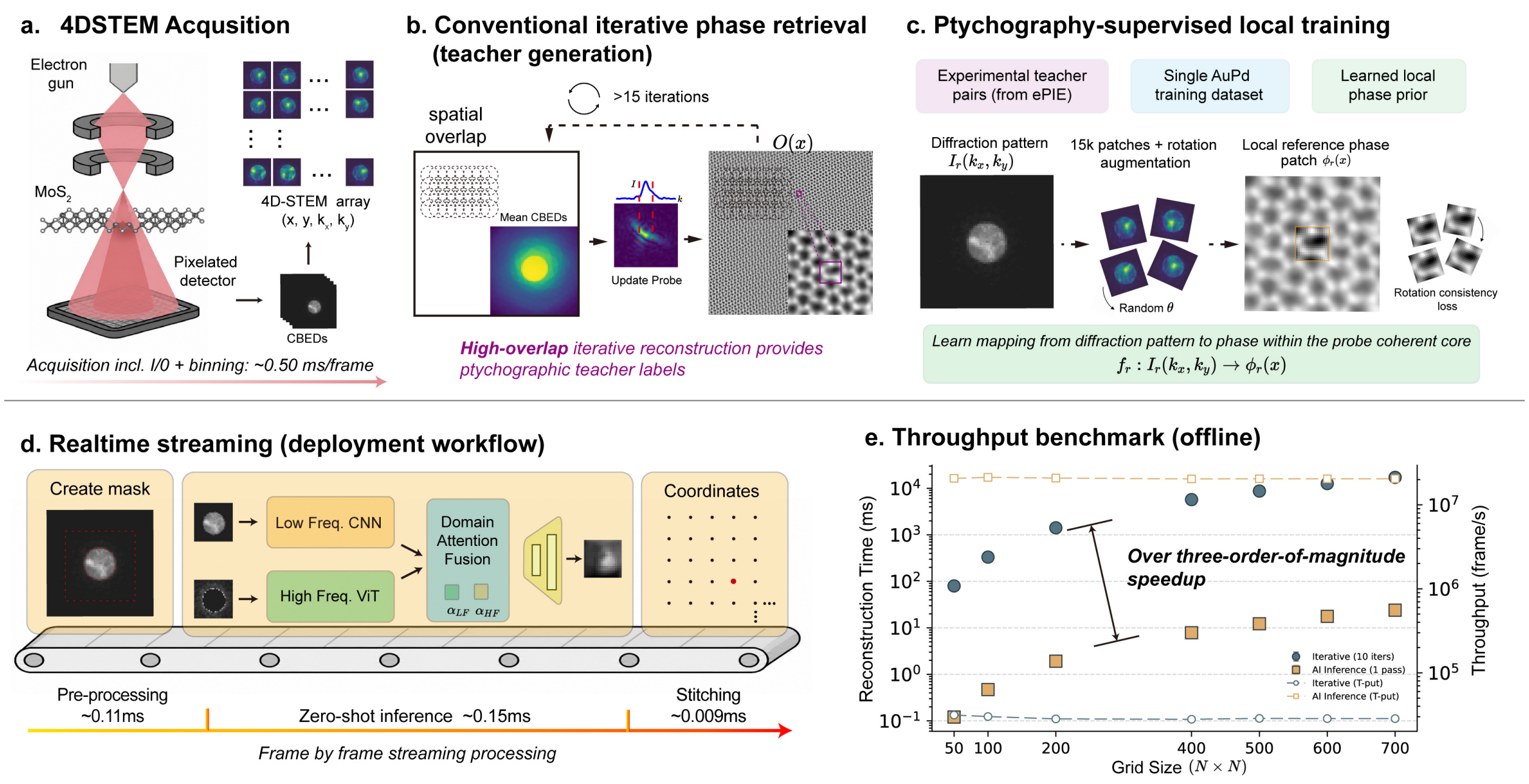

**Figure 1 | Ptychography-supervised local inference for real-time electron phase imaging.** (a) Experimental 4D-STEM acquisition. A focused electron probe is scanned across the specimen, and a pixelated detector records a CBED pattern at each probe position, forming a 4D-STEM datacube ($(x,y,k_x,k_y)$). The effective acquisition interval, including I/O and binning, is approximately 0.50 ms per frame.
(b) Conventional iterative phase retrieval for teacher generation. High-overlap 4D-STEM measurements are reconstructed by ePIE to generate coordinate-consistent ptychographic reference phase maps, which serve as physics-constrained teacher labels for local learning.

(c) Ptychography-supervised local training. Paired diffraction patterns and ePIE-derived local reference phase patches are extracted from a single experimental AuPd training dataset. Rotational augmentation and a rotation-consistency loss are used to stabilize the local diffraction-to-phase mapping within the probe coherent core.
(d) Real-time streaming deployment workflow. Each incoming diffraction frame is preprocessed, passed through the dual-branch local phase estimator, and assembled into a full-field phase image by deterministic stitching using the relative scan coordinates. The total online latency is approximately 0.27 ms per frame, including preprocessing, zero-shot inference, and stitching.
(e) Offline throughput benchmark across increasing scan-grid sizes. The one-pass inference pipeline achieves more than a three-order-of-magnitude speedup over iterative ePIE while maintaining stable high-throughput processing in batched mode.

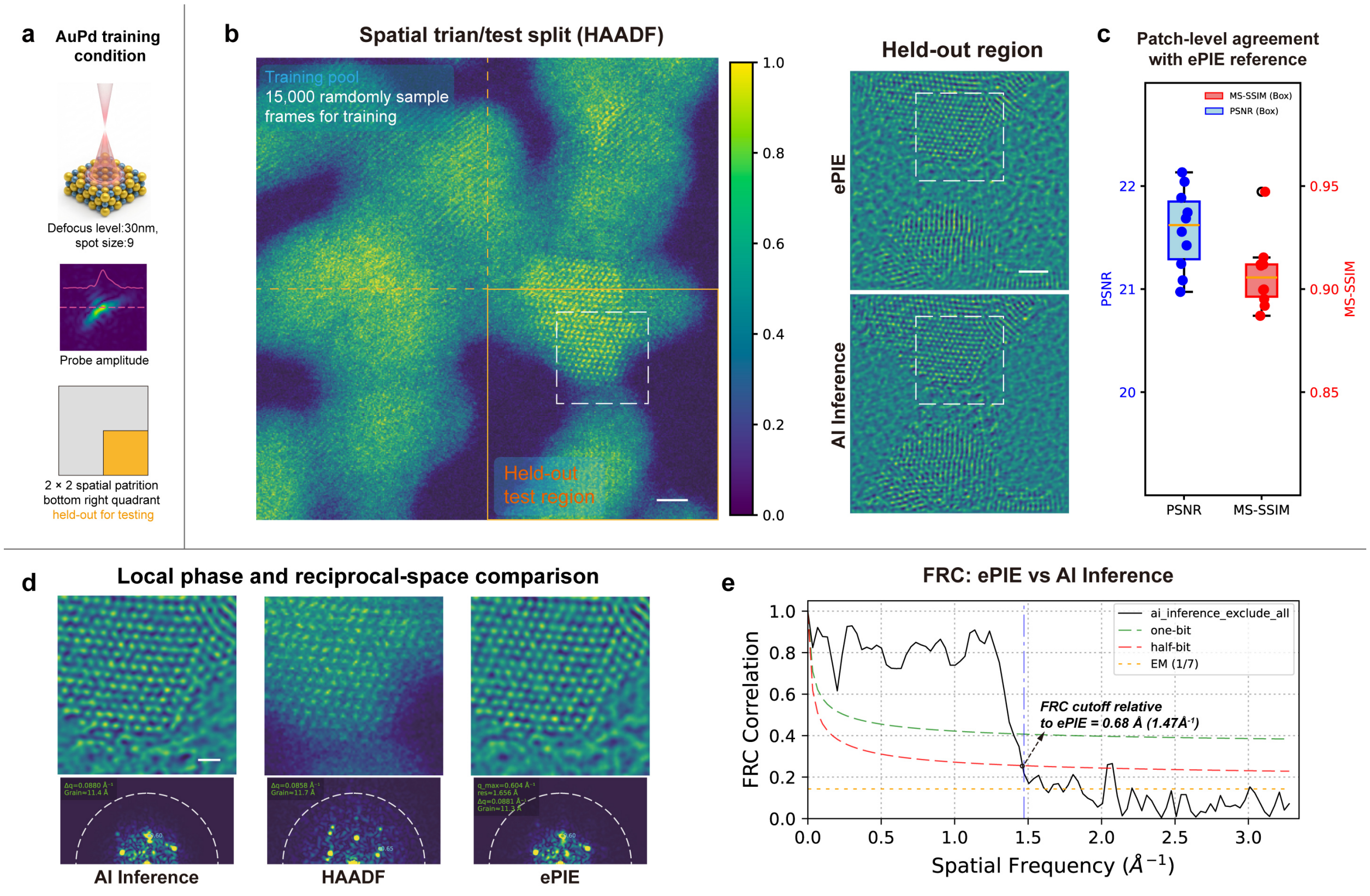

**Fig. 2 | Spatial hold-out validation on AuPd under matched imaging conditions.**
(a) AuPd training condition: spot size 9 and defocus (df = 30) nm. The corresponding probe amplitude profile and (2×2) spatial partition are shown, with the bottom-right quadrant reserved for testing.
(b) Spatial train/test split on the AuPd HAADF image. A total of 15,000 diffraction frames were randomly sampled from the training pool, while the bottom-right region was held out from training and used only for evaluation (scale bar, 10 Å). The right panels compare the ePIE reference and AI inference in the held-out region; dashed boxes mark the local region analyzed in (d) (scale bars, 3 Å).
(c) Patch-level agreement between AI inference and the ePIE reference in the held-out region,

summarized by PSNR and MS-SSIM box plots.
(d) Local real-space and reciprocal-space comparison for the representative held-out region. AI inference, HAADF, and ePIE reference images are shown with corresponding FFT magnitude maps (scale bars, 3 Å). The AI reconstruction preserves the dominant lattice periodicity and reciprocal-space peak structure of the ePIE reference.
(e) FRC between AI inference and the ePIE reference for the representative region in (d), yielding a cutoff of 0.68 Å relative to the ptychographic reference.

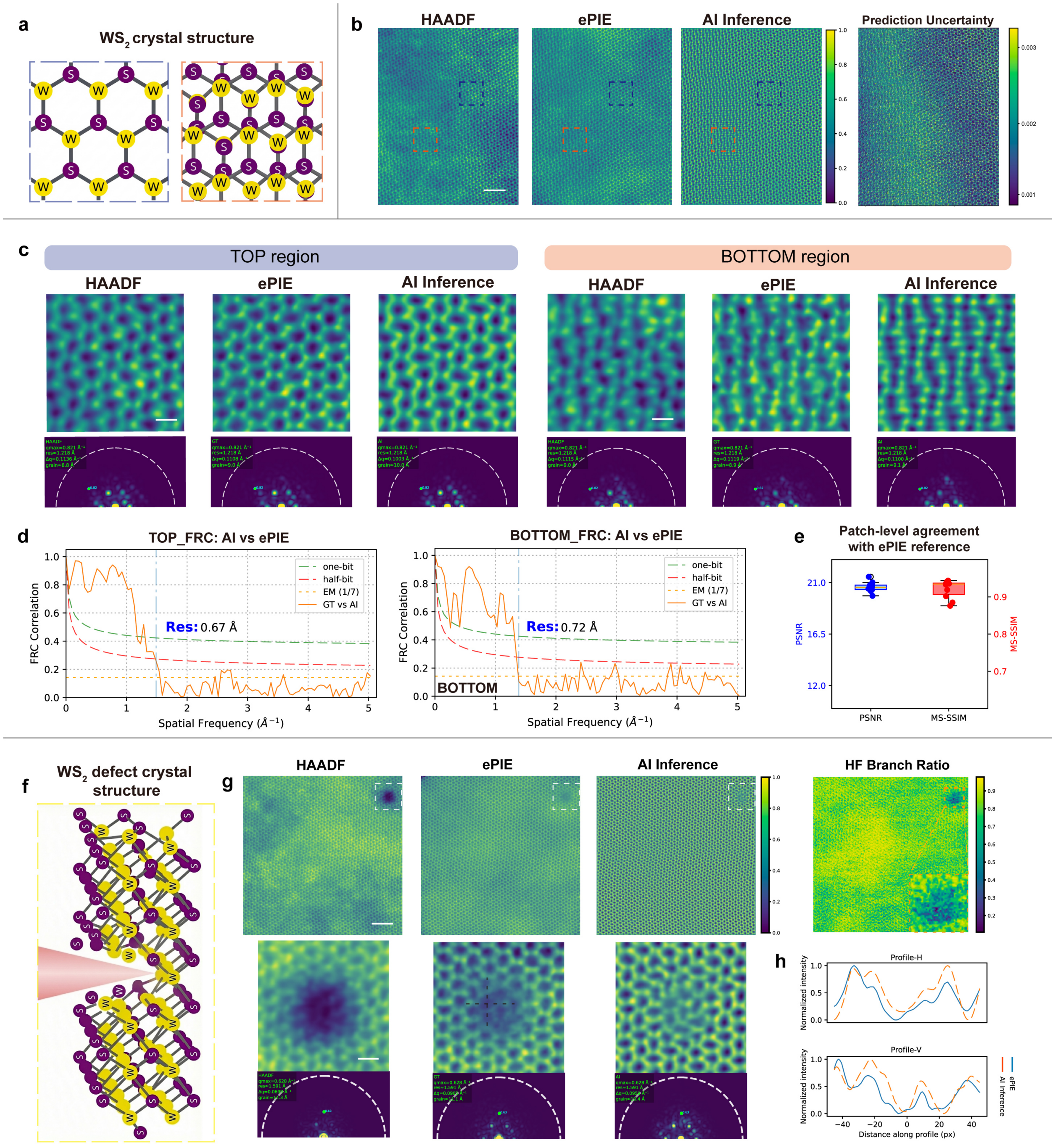

**Fig. 3 | Cross-material zero-shot transfer from AuPd to $WS_2$.**

(a) Atomic models of representative $WS_2$ regions, including a normal hexagonal lattice and a domain-overlap or bilayer-like structure.

(b) Full-field $WS_2$ reconstruction using the AuPd-trained model without fine-tuning. HAADF, ePIE reference, AI inference, and prediction uncertainty are shown for the same field of view (scale bars, 20 Å). Dashed boxes mark the TOP and BOTTOM regions analyzed in (c).

(c) Local comparison of the TOP and BOTTOM regions. HAADF, ePIE reference, and AI inference are shown with corresponding FFT magnitude maps (scale bars, 3 Å). The AI reconstruction preserves the dominant lattice periodicity and reciprocal-space peak structure of the ePIE reference.

(d) FRC between AI inference and the ePIE reference for the TOP and BOTTOM regions, yielding cutoffs of 0.67 Å and 0.72 Å, respectively.

(e) Patch-level agreement between AI inference and the ePIE reference over sampled $WS_2$ regions, summarized by PSNR and MS-SSIM.

(f) Schematic of the electron probe interacting with a $WS_2$ hole-like defect.

(g) Zero-shot reconstruction of the defect-containing region. Full-field HAADF, ePIE reference, AI inference, and the high-frequency branch ratio map are shown in the top row (scale bars, 20 Å), with enlarged defect-edge regions and FFT magnitude maps shown below (scale bars, 3 Å). The AI reconstruction follows the defect boundary while preserving the dominant lattice periodicity.

(h) Horizontal and vertical line profiles across the marked defect region, comparing AI inference with the ePIE reference.

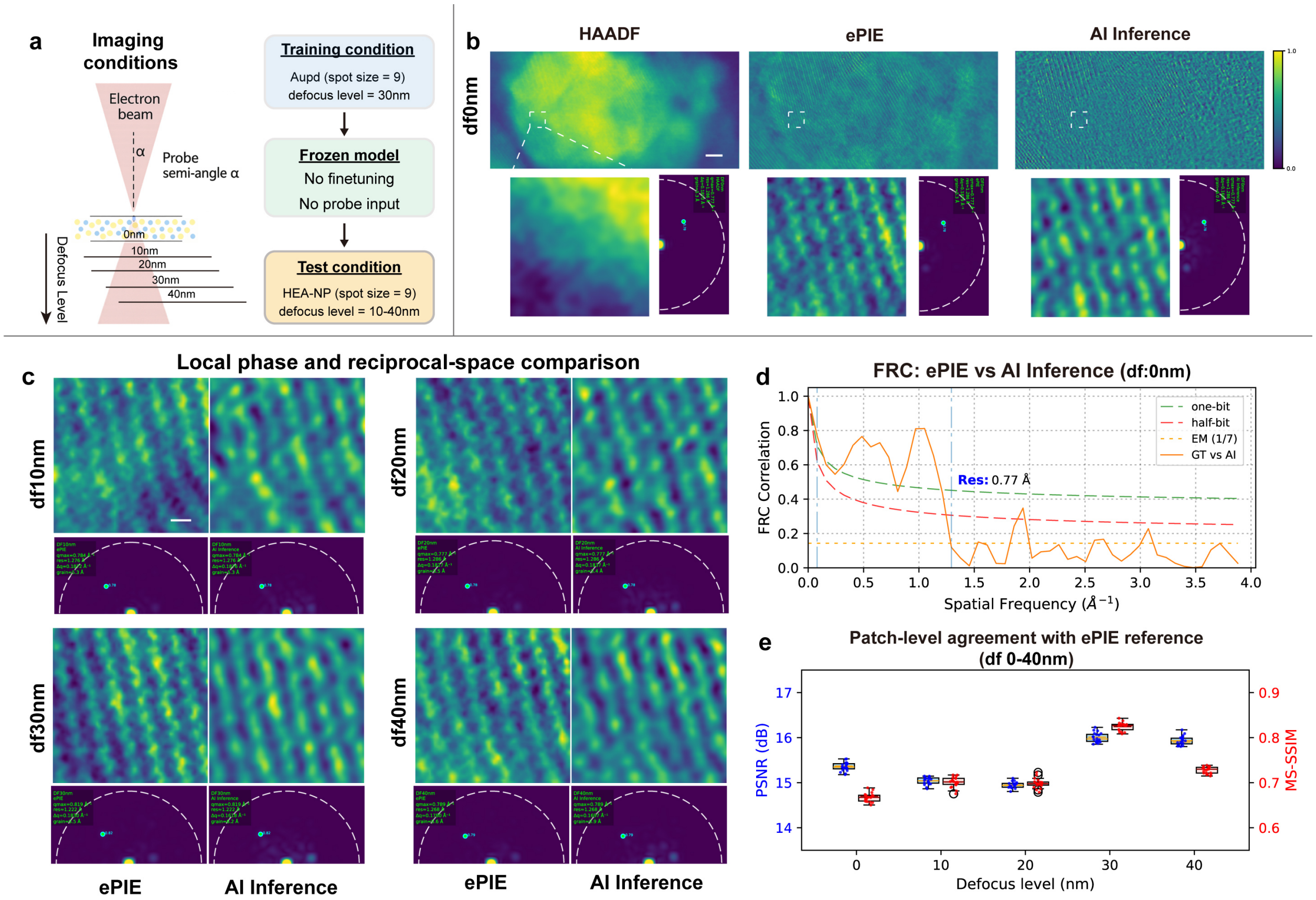

**Fig. 4 | Zero-shot phase inference under continuous defocus-induced transfer-function modulation.**

(a) Evaluation scheme for defocus-series testing. The model was trained only on AuPd acquired at spot size 9 and (df = 30) nm, then frozen and applied to HEA-NP datasets acquired at spot size 9 across (df = 0)–40 nm without fine-tuning or explicit probe input.

(b) Reconstruction at (df = 0) nm, representing the largest defocus shift from the training condition. HAADF, ePIE reference, and AI inference are shown with enlarged local regions and corresponding FFT magnitude maps (scale bars, 10 Å in the top row and 3 Å in the enlarged views).

(c) Local phase and reciprocal-space comparison across representative defocus levels. ePIE reference and AI inference are shown for (df = 10), 20, 30, and 40 nm, each with the corresponding FFT magnitude map (scale bars, 3 Å). The model preserves the main lattice contrast and reciprocal-space peak structure across the tested transfer conditions.

(d) FRC between AI inference and the ePIE reference at (df = 0) nm, yielding a cutoff of 0.77 Å relative to the ptychographic reference.

(e) Patch-level agreement with the ePIE reference across the full (df = 0)–40 nm series, summarized by PSNR and MS-SSIM. The best agreement occurs near the training defocus of (df = 30) nm, while structural consistency remains stable across the tested range.

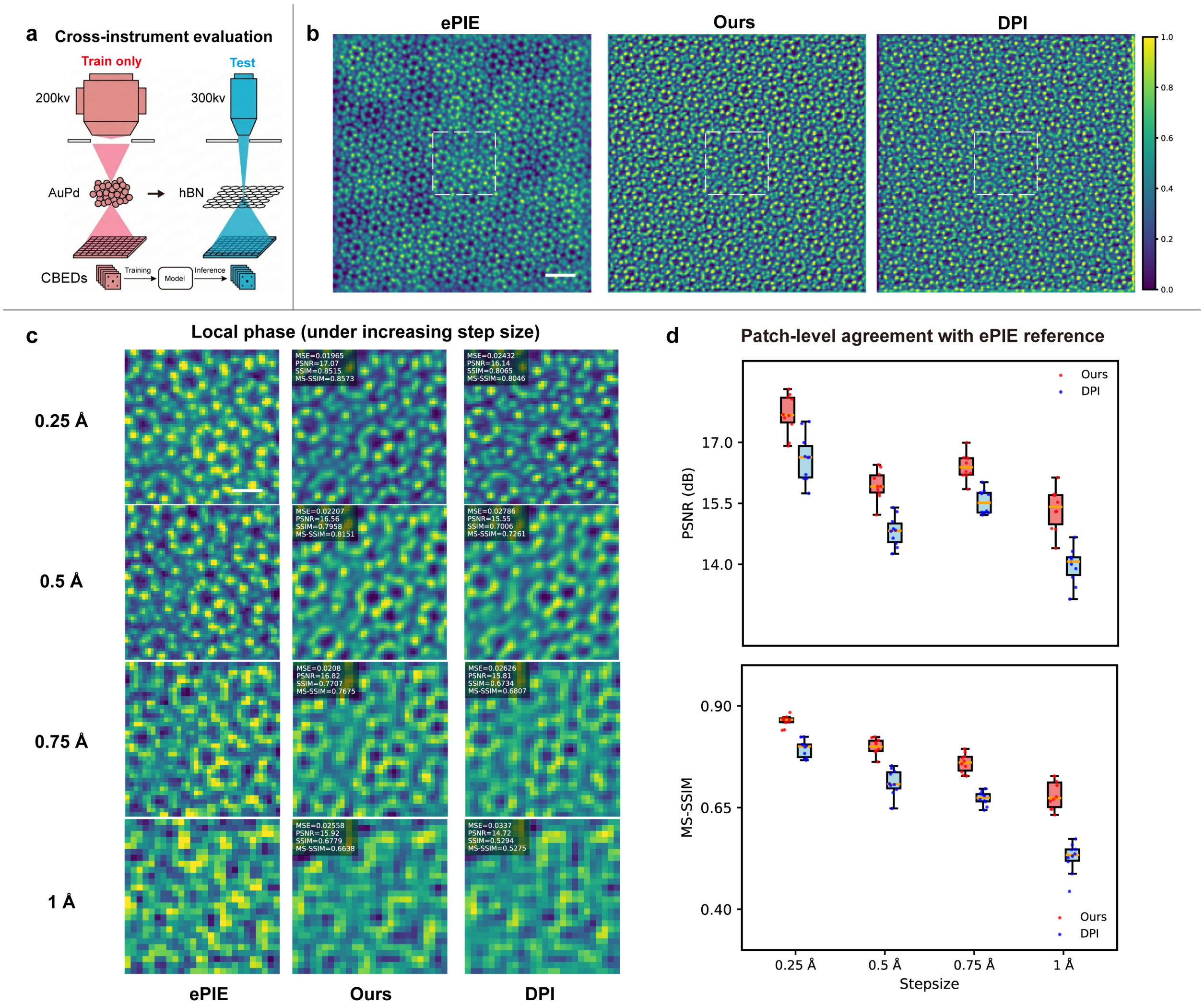

**Fig. 5 | Cross-instrument zero-shot transfer and sparse-sampling robustness.**
(a) Cross-instrument evaluation protocol. The model is trained only on experimental AuPd data acquired at 200 kV and applied directly to an hBN dataset acquired on a distinct 300 kV instrument, without fine-tuning or explicit probe input during inference.
(b) Full-field hBN reconstructions at a step size of 0.25 Å. ePIE reference, our zero-shot inference result, and the DPI baseline are shown for the same field of view (scale bars, 10 Å). The dashed boxes indicate the local region analyzed in (c).
(c) Local phase comparison under increasing scan step size from 0.25 to 1.0 Å. ePIE reference, our method, and DPI are shown for each sampling condition (scale bars, 3 Å). As the step size increases, our method retains clearer lattice contrast relative to DPI, although reduced overlap progressively limits full-field assembly.
(d) Patch-level agreement with the ePIE reference across different step sizes, summarized by PSNR and MS-SSIM box plots. Our method shows consistently higher agreement than DPI across the tested sampling densities.